\documentclass[runningheads]{llncs}
\usepackage[T1]{fontenc}
\usepackage{graphicx}
\begin{document}
\title{A wearable haptic device for edge and surface simulation}
% Third Author\inst{3}\orcidID{2222--3333-4444-5555} \and
% Third Author\inst{3}\orcidID{2222--3333-4444-5555} \and
% Third Author\inst{3}\orcidID{2222--3333-4444-5555} \and
% \email{lncs@springer.com}\\ \and

%%%%%%%%%%%%%%%%%%%%%%%%%%%%%%%%%%%

\author{Rui Chen\inst{1*}\orcidID{0000-0001-9178-9112} \and
Xianlong Mai\inst{1,2}\orcidID{0000-0001-8273-5112} \and
Alireza Sanaei\inst{1}\orcidID{0009-0004-3741-8846}\and
Domenico Chiaradia\inst{1}\orcidID{0000-0002-6183-0804} \and
Antonio Frisoli\inst{1}\orcidID{0000-0002-7126-4113}\and
Daniele Leonardis \inst{1}\orcidID{0000-0001-8612-7085} 
}
\authorrunning{R. Chen et al.}
\institute{Institute of Mechanical Intelligence, School of Advanced Studies Sant'Anna (SSSA), 56127 Pisa, Italy. \email{corresponding email: rui.chen@santannapisa.it} \and
CAS Key Laboratory of Mechanical Behavior and Design of Materials, Institute of Humanoid Robots, School of Engineering Sciences, University of Science and Technology of China, Hefei, China. \email{mxl2726@mail.ustc.edu.cn} }

\maketitle              % typeset the header of the contribution
\begin{abstract}
Object manipulation is fundamental to virtual reality (VR) applications, yet conventional fingertip haptic devices fail to render certain tactile features relevant for immersive and precise interactions, as i.e. detection of edges. This paper presents a compact, lightweight fingertip haptic device (24.3 g) that delivers distinguishable surface and edge contact feedback through a novel dual-motor mechanism. Pressure distribution characterization using a 6×6 flexible sensor array demonstrates distinct contact patterns between the two stimulation modes. A preliminary user study with five participants achieved 93\% average classification accuracy across four conditions (edge/surface contact with light/heavy pressure), with mean response times of 2.79 seconds. The results indicate that the proposed device can effectively convey edge and surface tactile cues, potentially enhancing object manipulation fidelity in VR environments.

\end{abstract}

\section{Introduction}

Virtual reality (VR) has become increasingly prevalent in applications ranging from gaming and training to teleoperation and medical simulation. However, the sense of presence and manipulation capability in VR environments remains limited if compared to the variety of haptic features perceived in natural tactile exploration \cite{frisoli2024wearable}. 
%While wearable haptic devices enable basic virtual hand control,  they cannot convey critical tactile information about contact location, contact geometry, or interaction forces. This limitation is particularly problematic during object manipulation tasks, where distinguishing between surface and edge contact is essential for stable grasping. When fingers contact an object's edge rather than its surface, the reduced contact area leads to decreased friction, potentially causing the object to slip or fall—yet current VR systems provide no indication of this critical transition.
Haptic technology offers a promising solution by delivering tactile feedback directly to users' fingers. Existing approaches primarily employ two modalities: normal force rendering \cite{PARK2018177,fluidreality,wolvering,Rui2025soft-thermal-haptic} and vibratory stimulation \cite{Piezoelectret,SAUVET2017100,Actuators_Review,HapThimble,science,FrontiersVR}. Normal force rendering devices provide intuitive feedback of grasping modulation, while vibratory actuators use transients and frequency variations as haptic cues for contact events.  To achieve richer haptic feedback, researchers have developed multi-DOF devices \cite{Daniele3DOF,KUANG2024103173,CHEN11007068,chinello2017three} that convey normal and tangential skin stretch, sliding, and surface orientation.  Although effective, these solutions typically exclude other geometric tactile features that can be relevant and informative in manipulation. Perception of object edges has shown to be crucial in tactile exploration \cite{bensmaia2008tactile,pruszynski2018fast}, evidencing fast and accurate edge orientation processing during object manipulation. On the other hand, rendering of edge contact by means of a wearable haptic device appears challenging, due to the multiple dofs involved for a generic, spatially located haptic feature with respect to the fingerpad. In this work, however, we propose to explore the topic through a feasible device design, implementing a limited number of dofs, yet effective to render the transition from a flat surface exploration to an edge. 
While previous work has investigated edge sharpness perception in real and virtual environments \cite{EdgeSharp2012,EdgeSharp2016,batik2025origami_shape_shifting} and developed edge displays for mobile devices \cite{MobileEdge,liu2025pneutouch}, these studies focused exclusively on edge rendering without addressing the transition between surface and edge contact, especially on the fingertip. We believe such transition is a fundamental requirement to provide informative tactile cues in object exploration and manipulation.

\begin{figure}
\includegraphics[width=\textwidth]{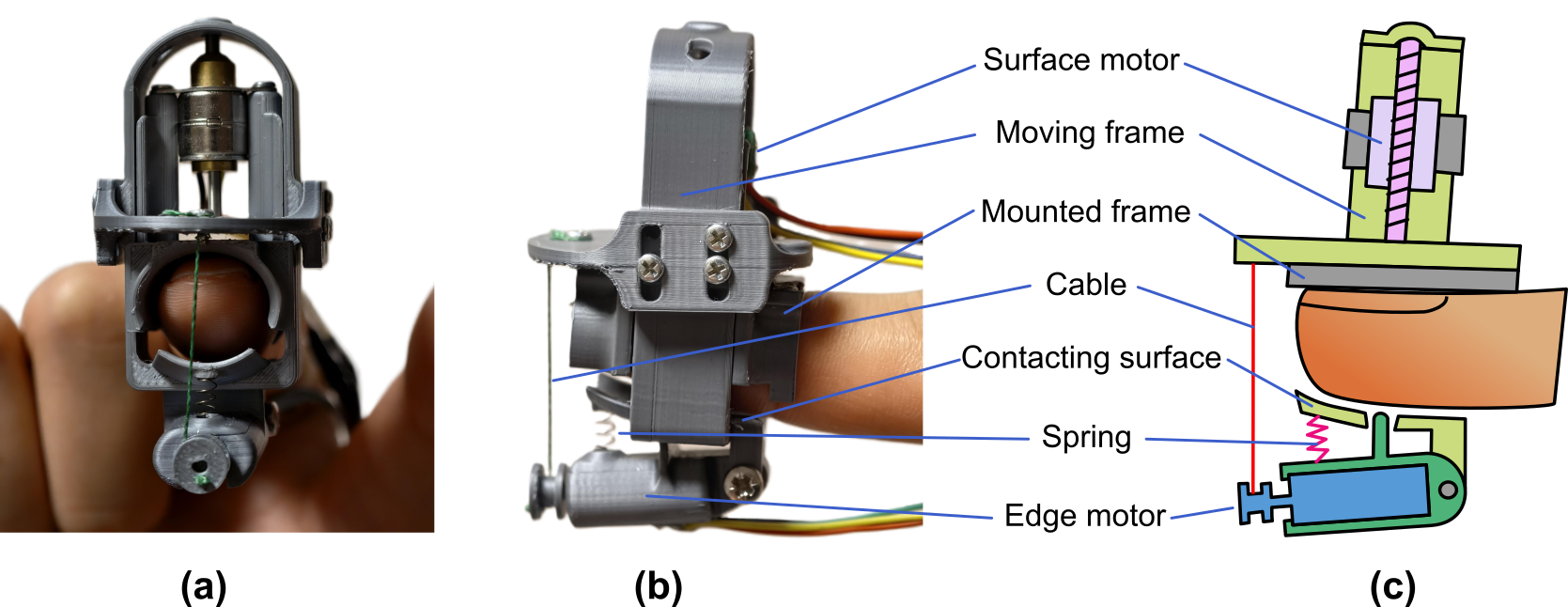}
\caption{Conceptual design of the haptic device showing the dual-motor mechanism for surface and edge contact stimulation.} \label{Conceptual}
\end{figure}

This paper presents a novel fingertip haptic device that addresses this gap by providing distinguishable surface and edge contact feedback through a compact dual-motor mechanism. The device weighs only 24.3 g and integrates with a portable wrist-worn control box (92 g), making it suitable for extended VR sessions. We characterize the pressure distribution patterns generated by the device using a 6×6 flexible sensor array and validate its perceptual distinguishability through a user study involving four tactile stimulation conditions. The proposed approach offers a foundation for more immersive and precise object manipulation in virtual environments.

\section{Methodology}

\subsection{Haptic Device Design}

\begin{figure}
\includegraphics[width=\textwidth]{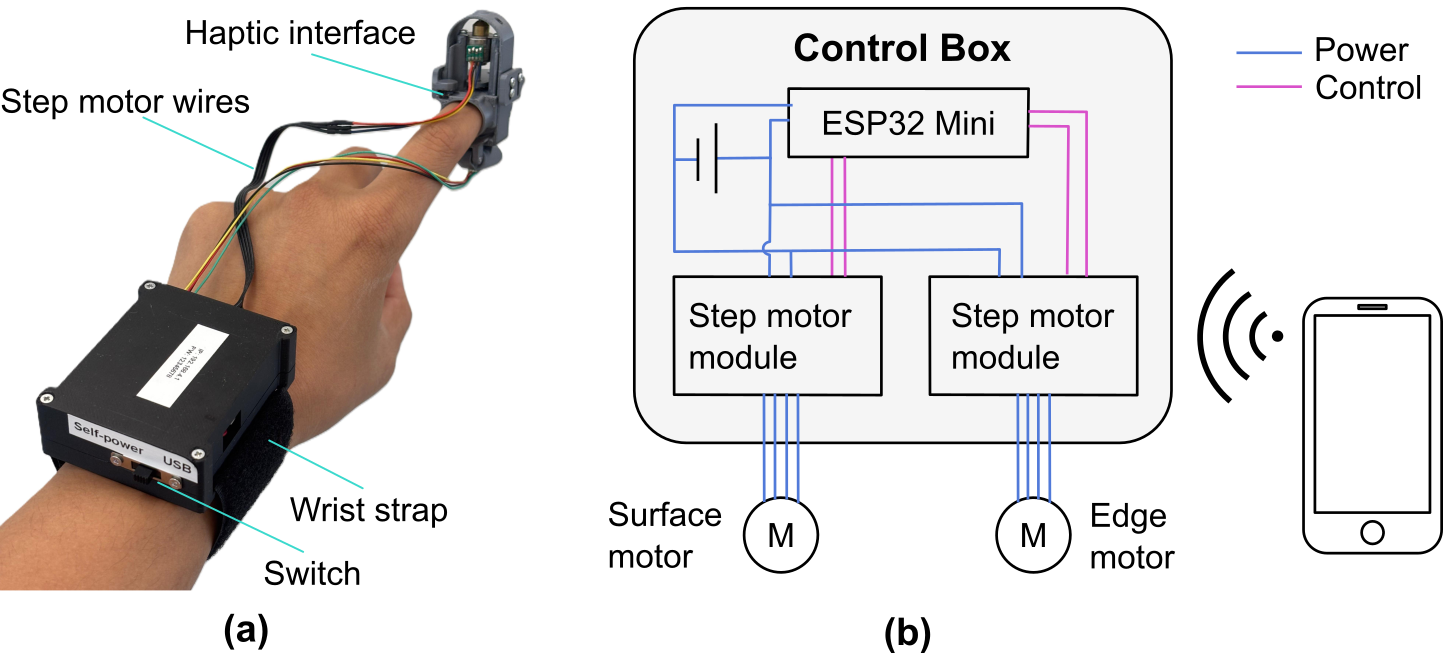}
\caption{(a) The wrist-mounted control box. (b) System architecture of the control box.} \label{Control}
\end{figure}

The fingertip haptic device comprises two independent actuation mechanisms for delivering surface and edge contact feedback (Fig. \ref{Conceptual}). The primary components include a surface motor to provide surface contact, an edge motor to provide edge contact, a moving frame with contacting surface and edge motor, a mounting frame for fingertip, a cable for edge motor, and a return spring to against the cable. The edge motor and cable are mounted on the moving frame, so the the edge is independent when the moving frame move.

\textbf{Surface Contact Mechanism:} A stepper motor with 8 mm body size and 6.5 mm stroke range controls the moving frame position. The motor features an 18° step angle (20 steps per revolution) and 0.3 mm stroke per rotation, providing 0.015 mm theoretical resolution and up to 3.18 ± 0.37 N output force. The motor mounts to a curved frame that fits comfortably on the finger. A curved contact surface attached to the moving frame delivers surface contact stimulation to the fingertip. By adjusting the moving frame position, the device modulates the contact force between the fingertip and contact surface.

\textbf{Edge Contact Mechanism:} The contact surface features a 2 mm wide elongated aperture through which an edge element can protrude to deliver edge contact stimulation. A 6 mm stepper motor (8° step angle, and 20 steps per revolution) with an integrated gearbox (26.45:1 ratio) controls the edge stimulus via a cable connected to the moving frame. Motor rotation drives the cable, regulating the edge protrusion height above the contact surface. The gearbox amplifies the output force by a factor of 26.45 and provides a 0.68° effective step angle. The motor delivers up to 2.21 ± 0.04 N cable tension, which is further amplified 2.63× through mechanical leverage, resulting in approximately 5.8 N maximum edge force. A passive return spring maintains the edge in a retracted position when the motor is inactive; however, this reduces the net deliverable edge force.

\textbf{Operation Modes:} For surface contact delivery, the surface motor positions the contact surface against the fingertip at the desired force level. For edge contact, two operational modes are available: in the first mode, the surface motor initially brings the contact surface near the fingertip, after which the edge motor tensions the cable to press the edge into the fingerpad. Alternatively, both the contact surface and edge can be actuated simultaneously, providing a more concentrated sensation at the edge. The compact design and lightweight stepper motors contribute to a total device weight of only 24.3 g.

\subsection{Portable Control System}

A wrist-mounted control box (Fig. \ref{Control}) manages device operation. The 92 g control box houses a custom PCB featuring an ESP32 microcontroller and two integrated stepper motor drivers. The ESP32 creates a local WiFi access point, enabling wireless control via smartphone or computer interface. A 3.7 V, 2000 mAh lithium battery provides power, with estimated consumption of 0.76 W and 0.52 W for the surface and edge motors respectively, supporting over 4 hours of continuous operation. In practice, intermittent motor usage extends operational time considerably.

\subsection{Pressure Distribution Characterization}

\begin{figure}
\includegraphics[width=\textwidth]{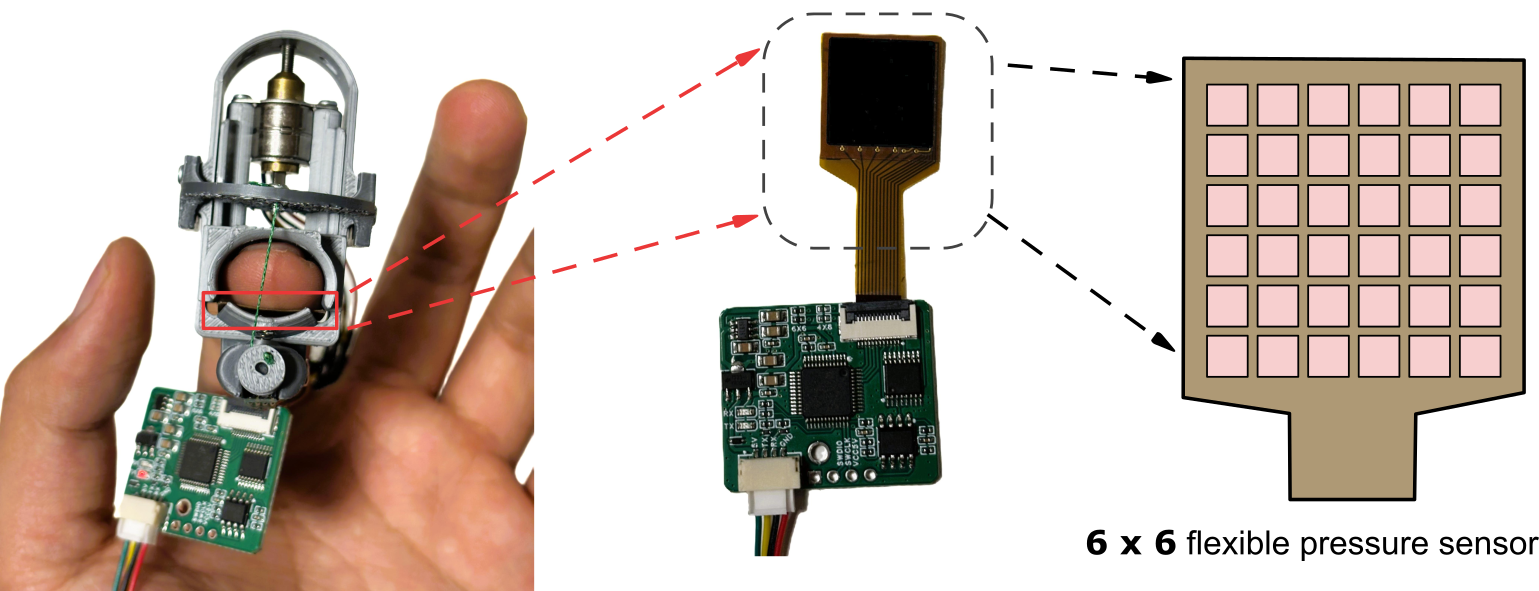}
\caption{Experimental setup for pressure distribution measurement using a 6×6 FSR array.} \label{HeatmapSetup}
\end{figure}

We characterized contact pressure distributions using a 6×6 flexible force-sensitive resistor (FSR) array placed between the fingertip and contact surface (Fig. \ref{HeatmapSetup}). The 36 independent sensors sampled pressure data at 10 Hz. During testing, both motors were commanded to maximum force for 3-4 seconds, and representative frames were selected for analysis.

\begin{figure}
\includegraphics[width=\textwidth]{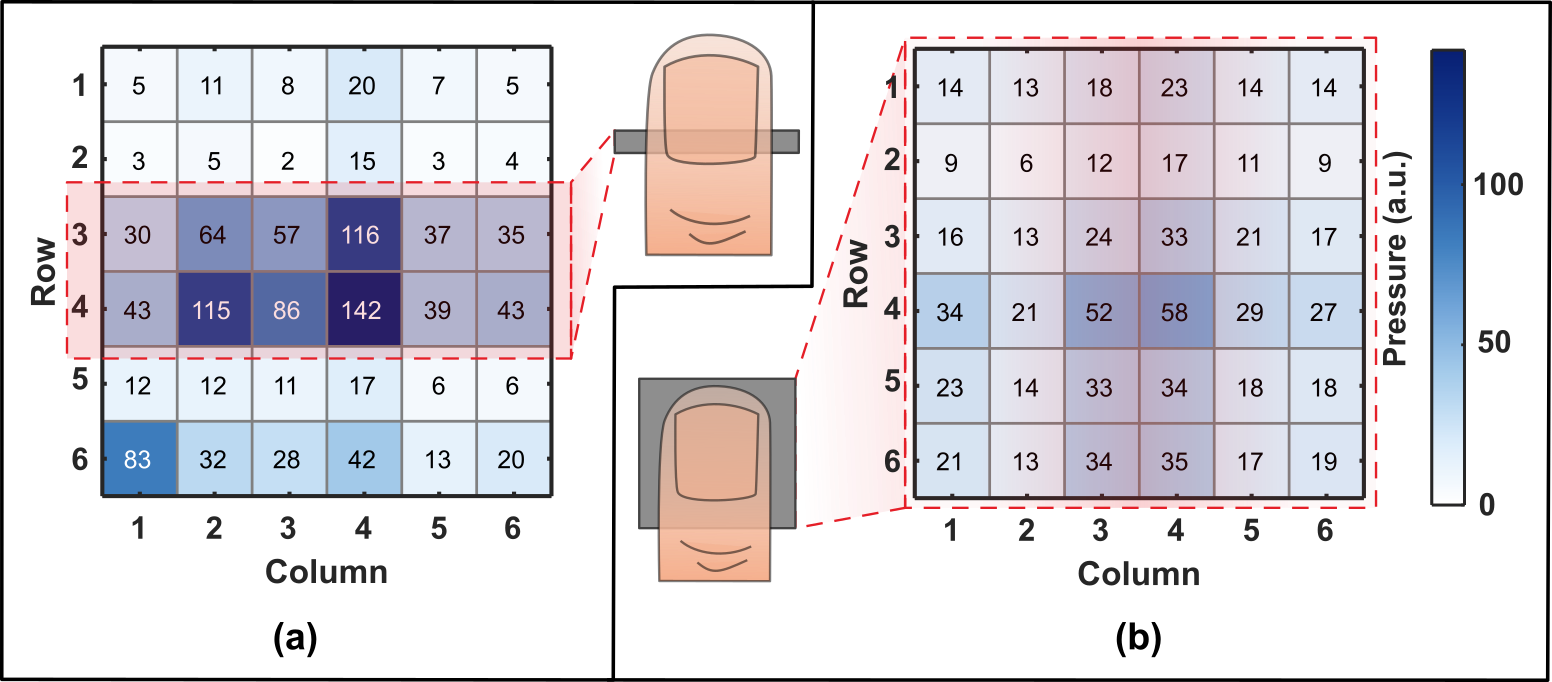}
\caption{Measured pressure distributions: (a) Edge contact showing concentrated linear pressure. (b) Surface contact showing distributed pressure pattern.} \label{Heatmap}
\end{figure}

Fig. \ref{Heatmap} shows the resulting pressure distributions. Edge contact produces a concentrated horizontal pressure band aligned with the physical edge position, while surface contact generates a broader distribution centered on the fingerpad. This clear distinction demonstrates the device's ability to deliver perceptually different tactile stimuli. Some elevated pressure values appear outside the primary edge contact region due to the FSR array's inherent stiffness and contact mechanics with the curved surface. Some outliers were observed in Fig.~\ref{Heatmap}, which resulted from unintended contact between the edge of the FSR array and the contacting surface due to slight mismatches in their shape and size.

\subsection{User Study}

\textbf{Participants:} Five male participants (mean age 30.4 ± 3.2 years, all right-handed) completed the study following approval by the Ethics Committee of Scuola Superiore Sant'Anna (protocol 412023). All participants provided informed consent.

\begin{figure}
\includegraphics[width=\textwidth]{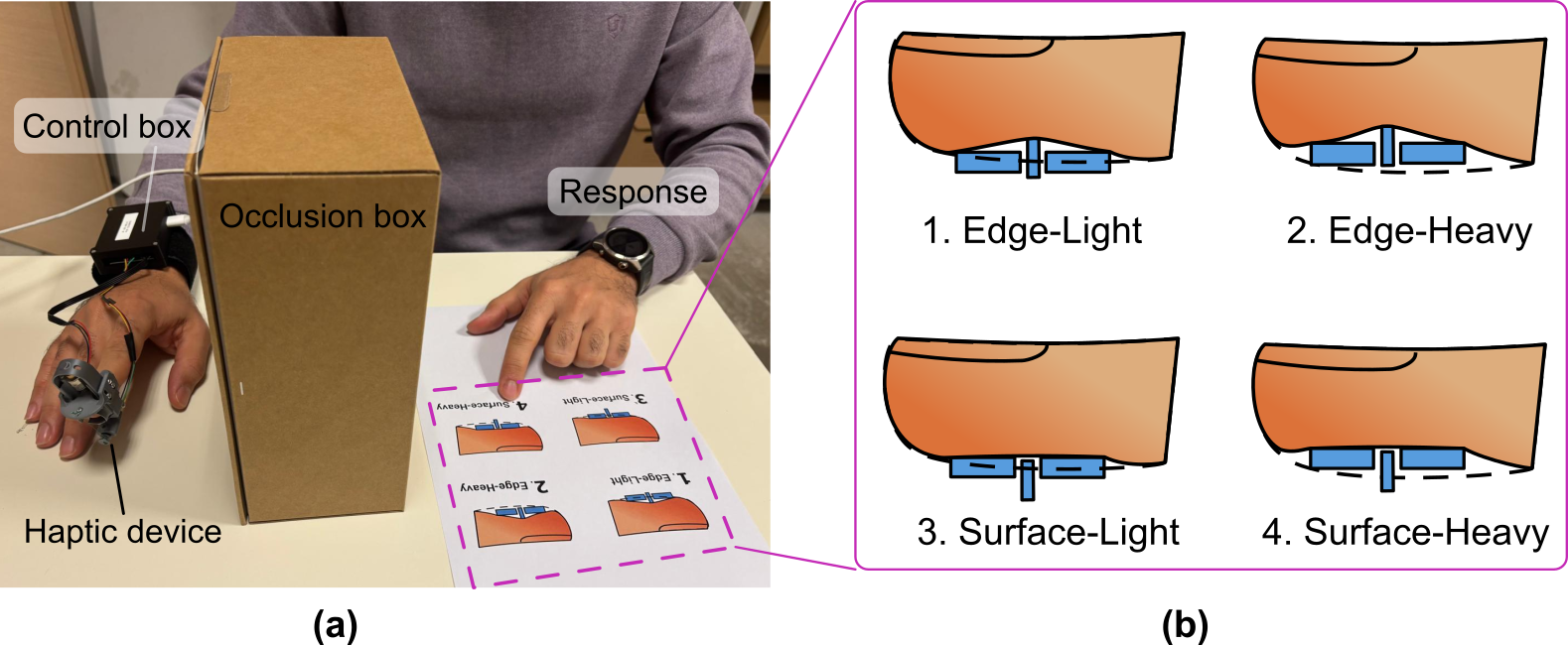}
\caption{User study protocol: (a) Experimental setup with participant wearing the device. (b) Stimulus presentations.} \label{SubjectSetup}

\end{figure}

\textbf{Experimental Conditions:} As shown in Fig. \ref{SubjectSetup}, four tactile stimulation conditions were tested: Edge-Light (EL), Edge-Heavy (EH), Surface-Light (SL), and Surface-Heavy (SH). Before formal testing, participants familiarized themselves with the device and all conditions.

\textbf{Calibration:} We defined motor displacement parameters as $a$ = 0.35 mm for the surface motor and $b$ = 1.5 mm for the edge motor cable displacement. For each participant, calibration was performed as follows: the surface motor was first driven to its maximum force position, then retracted by $2a$ to establish the zero position. Similarly, edge motor calibration involved advancing to maximum cable tension, followed by retraction of $2b$. The no-contact position was defined at $-a$ for the surface motor and $-b$ for the edge motor. The motor positions for the four experimental conditions were: EL ($a$, $b$), EH ($2a$, $2b$), SL ($a$, $-b$), and SH ($2a$, $-b$). By modulating the displacements of the two motors across these conditions, we generated distinct force sensations on the fingertip. Both motors operated at 200 steps/second, with the edge motor's effective speed reduced by its gearbox.

\textbf{Protocol:} Participants completed 20 trials (5 repetitions × 4 conditions) in randomized order while blindfolded. After each stimulus presentation, the device returned to the no-contact position for a 3-second inter-stimulus interval. Participants verbally reported their perceived condition, and responses were recorded along with response times.

\begin{figure}
\includegraphics[width=\textwidth]{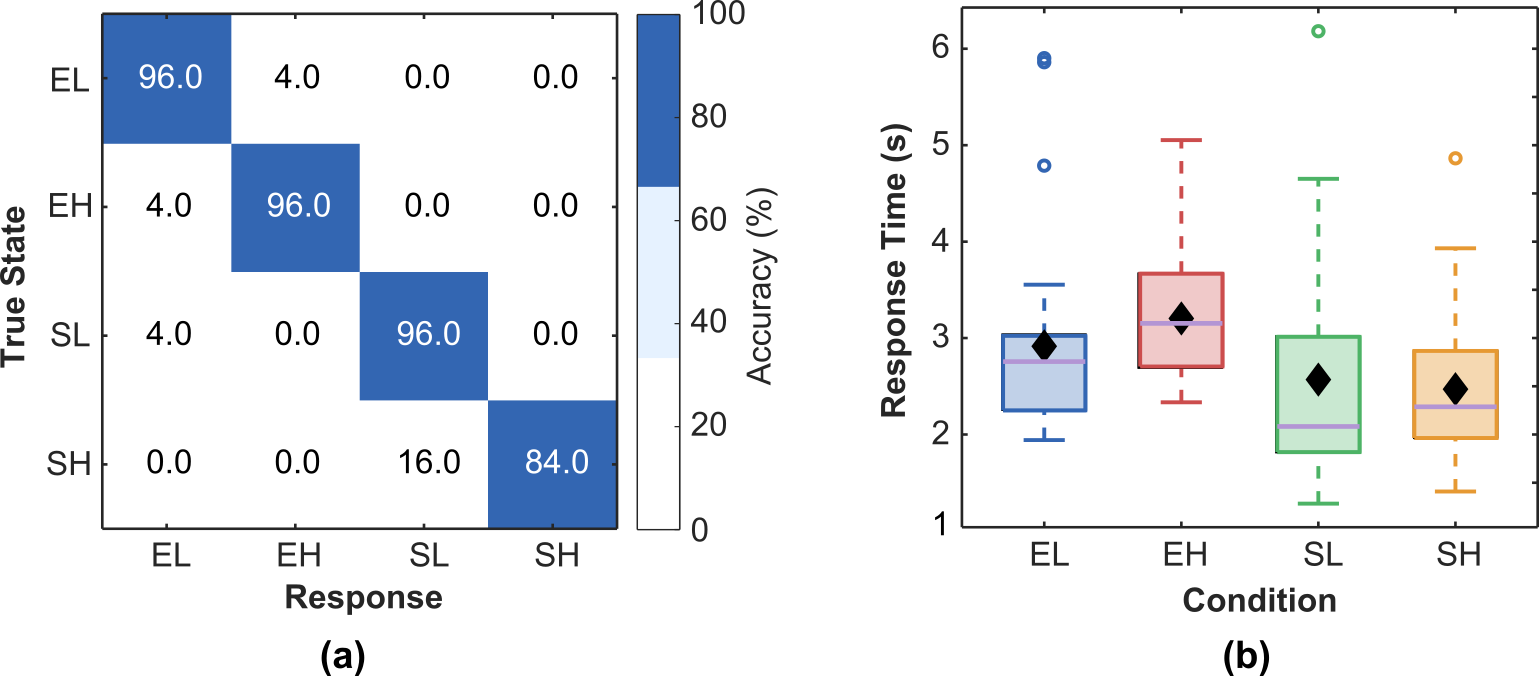}
\caption{User study results: (a) Classification accuracy across four conditions showing consistently high performance. (b) Response time distribution indicating rapid stimulus discrimination.} \label{Result}
\end{figure}

\textbf{Results:} Fig. \ref{Result} shows classification accuracy and response times. Mean accuracies were 96\% (EL), 96\% (EH), 96\% (SL), and 84\% (SH), with an overall accuracy of 93\%. Mean response times were 2.91 s (EL), 3.20 s (EH), 2.57 s (SL), and 2.47 s (SH), with an overall mean of 2.79 s. Response times include device actuation, perception, decision-making, and verbal response recording.

\section{Discussion and Conclusion}

This work presents a novel fingertip haptic device capable of delivering distinguishable surface and edge contact stimuli through a compact dual-motor mechanism. Pressure distribution measurements confirmed distinct tactile patterns between the two contact modes, and user evaluation demonstrated 93\% classification accuracy with mean response times under 3 seconds. These results indicate that the device can effectively convey critical contact geometry information for object manipulation in VR environments.

The device's compact form factor (24.3 g) and wrist-mounted control box (92 g) make it suitable for extended use in VR applications. The high classification accuracy across light and heavy pressure conditions suggests robust perceptual discriminability, while the rapid response times indicate potential for real-time VR interaction. The distinguishable edge and surface feedback addresses a fundamental limitation of current VR haptic systems, potentially enabling more natural and precise virtual object manipulation.

Several limitations warrant consideration. First, while the ideal rendering of edges involves multiple DoFs, we focused on a feasible fingertip haptic design for rendering flat surface-to-edge transition, which can be crucial for preliminary investigation of edge information in virtual manipulation. Future designs might consider DoFs related to edge orientation and translation with respect to the fingerpad. Second, stepper motors enable position control within the operational force range; however, accurate force estimation remains challenging. Integrating force sensors would enable closed-loop force control, allowing real-time adaptation to individual fingertip compliance. Third, the preliminary study involved only five male participants, and reaction times were recorded via verbal reporting, which introduces articulatory latency. A larger, gender-balanced sample with a digital response interface (e.g., button press or touchscreen GUI) would improve both result reliability and measurement precision. Fourth, acoustic isolation was not implemented during the experiments; motor noise could have served as unintended cues, which will be addressed with noise-cancelling headphones in future studies. Fifth, the 2~mm wide aperture may affect surface perception, warranting further investigation. While the 24.3~g weight is acceptable for VR applications, alternative actuation methods such as fabric-based pneumatic actuators \cite{Rui2025Glove,Rui2025LPPAMs} may provide lighter wearability. Finally, long-term durability testing and design optimization (e.g., reducing motor noise, improving ergonomics) require further work.
 
Future work will focus on integrating the device into complete VR manipulation scenarios (e.g., virtual object assembly and robotic teleoperation), extending the device to multiple fingers for whole-hand haptic feedback, conducting a comprehensive user study with a larger and gender-balanced cohort using standardized psychophysical protocols (digital response interface and acoustic isolation), and exploring alternative actuation systems for richer feedback modalities. The proposed approach provides a starting point for the challenging topic of edge rendering with wearable haptics, aiming at next-generation VR haptic interfaces that convey comprehensive tactile information about contact geometry and interaction forces.

\bibliographystyle{splncs04}
\bibliography{mybib}




% LNCS Homepage, \url{http://www.springer.com/lncs}, last accessed 2023/10/25
% \end{thebibliography}
\end{document}